\documentclass[conference]{IEEEtran}
\IEEEoverridecommandlockouts

\usepackage{amsmath,amssymb,amsfonts}
\usepackage{algorithmic}
\usepackage{graphicx}
\usepackage{textcomp}
\usepackage{xcolor}
\usepackage{bm}
\usepackage{subfigure}
\usepackage{hyperref}
\usepackage{tikz}

\def\BibTeX{{\rm B\kern-.05em{\sc i\kern-.025em b}\kern-.08em
    T\kern-.1667em\lower.7ex\hbox{E}\kern-.125emX}}
\begin{document}

\title{ Bi-Level Spatial and Channel-aware Transformer for Learned Image Compression 
}

\author{\IEEEauthorblockN{Hamidreza Soltani} \IEEEauthorblockA{\textit{Department of Electrical Engineering   }\\
\textit{Tarbiat Modares University}\\
 \href{mailto:hamidrezasoltani@modares.ac.ir}{hamidrezasoltani@modares.ac.ir}}
 \and
 \IEEEauthorblockN{Erfan Ghasemi}
 \IEEEauthorblockA{\textit{Department of Electrical Engineering }\\
 \textit{Tarbiat Modares University} \\
 \href{mailto:ghasemi.erfan@modares.ac.ir}{ghasemi.erfan@modares.ac.ir}}

\IEEEauthorblockA{\textit{}}
}

\maketitle

\begin{abstract}

Recent advancements in learned image compression (LIC) methods have demonstrated superior performance over traditional hand-crafted codecs. These learning-based methods often employ convolutional neural networks (CNNs) or Transformer-based architectures. However, these nonlinear approaches frequently overlook the frequency characteristics of images, which limits their compression efficiency. To address this issue, we propose a novel Transformer-based image compression method that enhances the transformation stage by considering frequency components within the feature map. Our method integrates a novel Hybrid Spatial-Channel Attention Transformer Block (HSCATB), where a spatial-based branch independently handles high and low frequencies at the attention layer, and a Channel-aware Self-Attention (CaSA) module captures information across channels, significantly improving compression performance. Additionally, we introduce a Mixed Local-Global Feed Forward Network (MLGFFN) within the Transformer block to enhance the extraction of diverse and rich information, which is crucial for effective compression. These innovations collectively improve the transformation's ability to project data into a more decorrelated latent space, thereby boosting overall compression efficiency. Experimental results demonstrate that our framework surpasses state-of-the-art LIC methods in rate-distortion performance.

\end{abstract}


\section{\textbf{Introduction}}

Lossy image compression is a crucial area of research in signal processing and computer vision, widely used to decrease storage and transmission costs. Over the past decades, numerous traditional standards like JPEG \cite{jpeg}, JPEG2000 \cite{jpeg2000}, and VVC \cite{vtm2022} have been developed, incorporating manually designed modules to enhance compression performance. On the other hand, with rapid advancements in deep learning, end-to-end learned lossy image compression has significantly progressed in recent years. Notably, some recent learning-based image compression methods \cite{chen2022two, he2022elic, wang2022neural, zhu2022transformer} have even surpassed VVC, the leading traditional codec.

Transform coding is the fundamental technique used in image compression \cite{yang2022introduction}. Codecs based on transform coding divide the lossy compression process into three key stages: transformation, quantization, and entropy coding. In neural image compression, each of these stages can be effectively performed using deep neural networks. The transformation stage typically employs convolutional neural networks (CNNs), which excel at learning nonlinear functions \cite{leshno1993multilayer}. These nonlinear functions have the capability to map data into a latent space, facilitating more efficient compression compared to the linear transforms used in traditional codecs. The entropy model employs deep generative models to estimate the probability distribution of the latent representation \cite{balle2018a,minnen2018joint, he2021checkerboard}. Furthermore, various differentiable quantization methods \cite{balle2016end, agustsson2017soft,yang2020variational} are introduced to enable end-to-end training of the entire neural compression framework. It is apparent that any improvement in these three stages of the data compression algorithm can lead to increased compression efficiency.

\begin{figure*}[tp]
    \centering
    \includegraphics[width=1\linewidth]{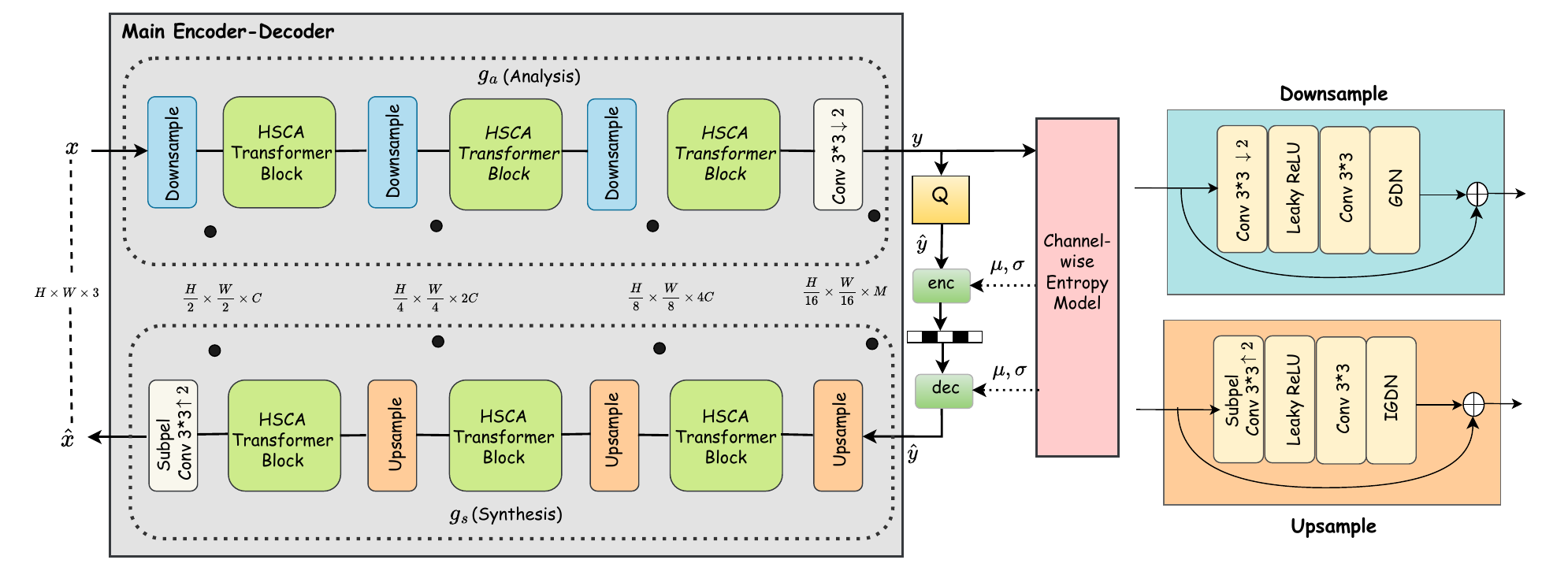}
    \caption{ Left: This illustrates the overall framework of our proposed neural image compression model. The model utilizes multiple downsample blocks, upsample blocks, and Hybrid Spatial-Channel Attention Transformer Blocks (HSCATB) to build the nonlinear transforms (i.e., analysis transform $g_a(.)$ and synthesis transform $g_s(.)$). Right: This shows the architectures of the downsample and upsample blocks.}
 
    \label{fig:visual-comparison}
\end{figure*}

Most learned image compression methods usually leverage convolutional neural networks (CNNs) to achieve nonlinear transformation \cite{fu2021learned, balle2016end, minnen2018joint}. However, the local receptive fields of CNNs restrict their ability to represent information effectively and result in redundant latent representations. To address the issue of capturing non-local information with CNN-based approaches, recent research has turned to attention modules or Transformers \cite{chen2019neural,zou2022devil, zhu2022transformer}, which can capture global spatial relationships and thus improve compression efficiency.

Despite the success of Transformer-based neural compression approaches, they often overlook the frequency characteristics of images during compression. To address this limitation, we propose a novel Transformer-based image compression method that enhances the transformation stage by considering the frequency components of the feature map. Our approach integrates a spatial-aware self-attention (SaSA) module, allowing for the independent handling of high and low frequencies within a feature map at the attention layer, significantly improving overall compression efficiency. In addition, our Transformer block takes advantage of a channel-aware attention (CaSA) module, which is capable of extracting channel-wise global information. Furthermore, We augment our Transformer block with a feature enhancement feed-forward network, called the Mixed Local-Global Feed Forward Network (MLGFFN), to enhance the extraction of diverse and rich information, which is crucial for compression tasks. These enhancements collectively aim to boost the transformation's ability to project data into a more decorrelated latent space.


The structure of the paper is as follows: Section II offers an in-depth review of learned image compression techniques and vision Transformers. Section III outlines the problem formulation and provides a detailed description of our proposed method. Section IV discusses the implementation details and presents the experimental results. Finally, Section V concludes the paper.

\section{\textbf{Related Work}}

\subsection{\textbf{Learned Image Compression (LIC)}}

Neural image compression methods utilize a nonlinear transform coding paradigm that resembles variational autoencoder (VAE) models \cite{goyal2001theoretical}. {Ball{{e}} \emph{et al.} \cite{balle2017endtoend} were the first to model the image compression framework as a compressive autoencoder, achieving results comparable to the JPEG2000 \cite{jpeg2000} standard. They introduced a fully factorized entropy model to estimate the distribution of the latent representation for computing the bit-rate. To achieve a more accurate entropy model, authors in \cite{balle2018a} proposed a hyperprior entropy model. Furthermore, Minnen \emph{et al.} \cite{minnen2018joint} integrated a local context block with a hyperprior network as an entropy model to improve image compression performance. Alongside the local context, Qian \emph{et al.}  \cite{qian2020learning} introduce a global context model within the entropy model to leverage long-range correlations. In a different approach \cite{minnen2020channel}, channel-wise context modeling was implemented instead of spatial context to speed up decoding times. Zafari \emph{et al.} \cite{zafari2023frequency} introduced an innovative feature-level frequency disentanglement approach that reduces the bitrate by enabling more precise quantization, thereby improving rate-distortion performance. some works \cite{agustsson2023multi, mentzer2020high,yang2024lossy} focus on enhancing the realism of reconstructed images in the image compression pipeline. Mentzer \emph{et al.} \cite{mentzer2020high} integrated a conditional GAN into an autoencoder-based compression model to improve perceptual quality of decoded image. More recently, Agustsson \emph{et al.} \cite{agustsson2023multi} introduced a Multi-Realism model utilizing PatchGAN \cite{chang2019free}, which outperforms HiFiC \cite{mentzer2020high}.

\subsection{\textbf{Vision Transformers (ViTs)}}

Inspired by the effectiveness of Transformers \cite{vaswani2017attention} in natural language processing, Vision Transformers (ViTs) \cite{dosovitskiy2020image} were pioneers in introducing Transformers to visual tasks. They achieved this by employing the self-attention (SA) mechanism on non-overlapping patches. Vision transformers demonstrate remarkable performance across a wide range of tasks, excelling not only in high-level vision tasks like image classification \cite{touvron2021training,khoshkhahtinat2023dynamic} but also in low-level tasks such as image restoration \cite{liang2021swinir, wang2022uformer}, image denoising \cite{zhang2023xformer}, and learned image compression \cite{zhu2022transformer,zou2022devil,khoshkhahtinat2023multi,qian2022entroformer,liu2023learned}. In Transformer-based image compression research, these studies \cite{zou2022devil, zhu2022transformer} utilized the Swin Transformer \cite{liu2021swin} to create a nonlinear transformation, demonstrating its ability to achieve higher compression efficiency when compared to convolution-based models. Qian \emph{et al.}  \cite{qian2022entroformer} proposed a Transformer-based entropy model to capture global context. The authors in \cite{khoshkhahtinat2023multi} developed a primary autoencoder based on a novel Transformer architecture, which aims to effectively extract both local and global information while also reducing computational complexity. Recent research \cite{liu2023learned} combined the CNN's local capability with the Transformer's non-local ability to improve compression performance . 


\begin{figure*}[tp]
    \centering
    \includegraphics[width=.99\linewidth]{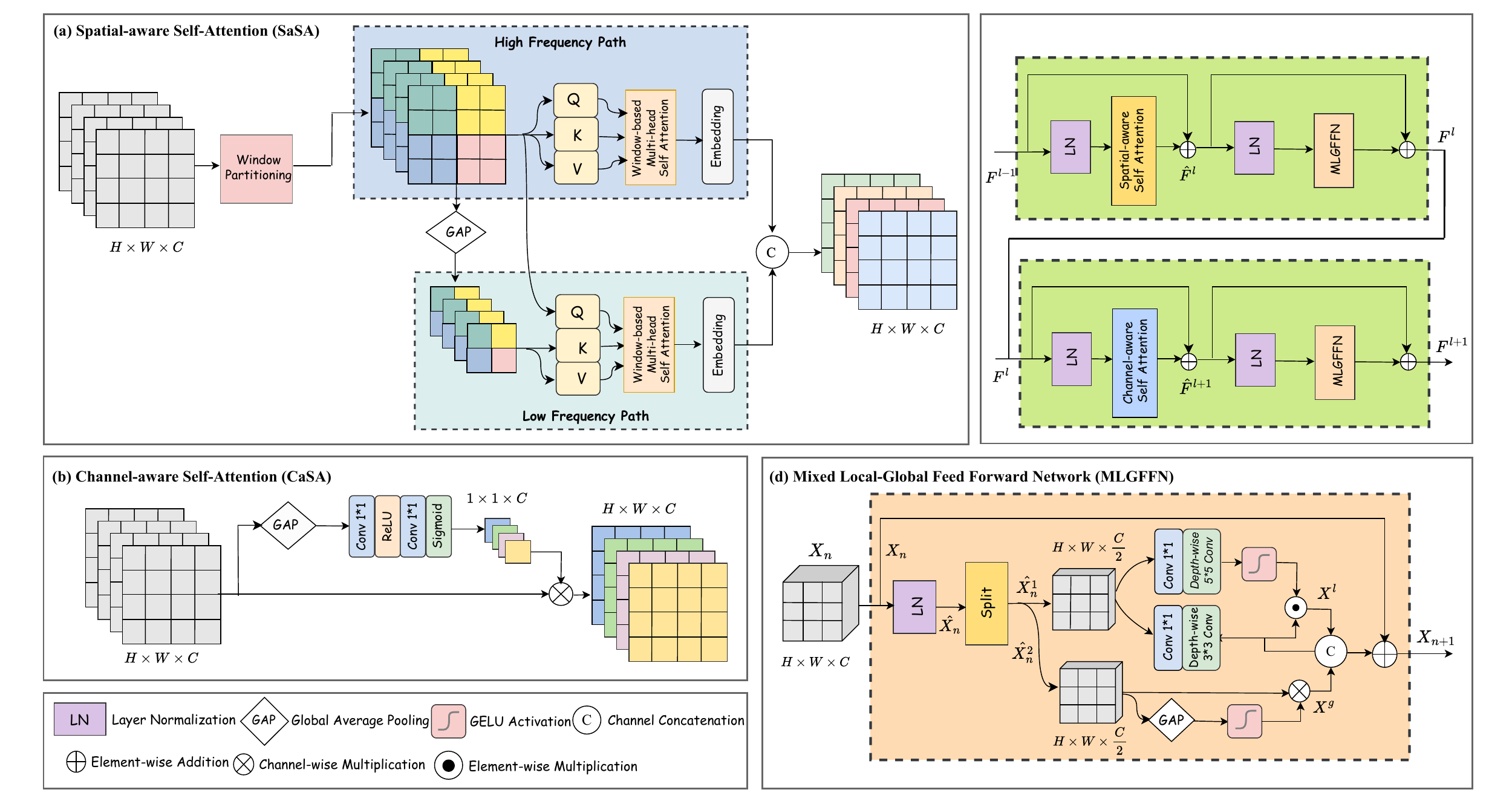}
    \caption{ (a) Spatial-aware Self-Attention (SaSA), which is composed of low-frequency and high-frequency paths. (b) Channel-aware Self-Attention (CaSA). (c) Hybrid Spatial- Channel Attention Transformer Block
(HSCATB). (d) Mixed Local-Global Feed Forward Network (MLGFFN). }
 
    \label{fig:visual-comparison}
\end{figure*}

\section{\textbf{Methods}}

\subsection{\textbf{Problem Formulation}}

In learned image compression, the network can be trained with a loss function that balances between rate and distortion. Thus, the rate-distortion optimization $L$ can be expressed as follows:
\begin{equation}
    L= R+\lambda D
    \label{eq:rate-distortion},
\end{equation}

where $D$ represents the distortion between the input image and the reconstructed image, which can be measured by distortion metrics such as mean squared error (MSE). $R$ denotes the rate term, which represents the file size of the compressed image.

Most learning-based image compression approaches are founded on the variational autoencoder (VAE) \cite{kingma2013auto}. These neural compression frameworks consist of an analysis transform, quantization, synthesis transform, and an entropy model used to estimate bit-rates \cite{goyal2001theoretical}. In the learned compression approaches, initially, at the encoder side, the input image $\bm{x}$ undergoes an analysis transformation $g_a(.)$ to generate a decorrelated latent representation. Subsequently, to decrease the number of bits required for image storage or transmission, the continuous latent representation $\bm{y}$ is discretized and then compressed losslessly using entropy coding methods with an estimated distribution of the quantized latent representation. Finally, at the decoder side $g_s(.)$, the received quantized latent representation $\bm{\hat{y}}$ is utilized to reconstruct the image through a synthesis transform \cite{balle2020nonlinear}. The above process can be formulated as:
\begin{equation}
 \begin{aligned}
    &\bm{y}= g_a(\bm{x};\bm{\phi})\\ &\bm{\hat{y}}=Q(\bm{y}),
    \\&\bm{\hat{x}}= g_s(\bm{\hat{y}};\bm{\theta}),
\end{aligned}
\end{equation}
where $\bm{\phi}$ and $\bm{\theta}$  denote the parameters of analysis transform and synthesis transform respectively which are learned jointly in optimzing compression network. $Q$ shows the quntization operation.

The loss for the rate term is defined as the cross-entropy between the true distribution and the estimated distribution of the latent representation. If the entropy model is constructed based on the hyper-prior \cite{balle2018a}, the bit rate can be expressed as:
\begin{equation}
 R= \bm{E}_{x\sim p_X}[-\log p_{\bm{\hat{y}}|{\bm{\hat{z}}}}(\bm{\hat{y}}|{\bm{\hat{z}}},\bm{\zeta})- \log p_{\bm{\hat{z}}}({\bm{\hat{z}}}, \bm{\psi})]. 
\label{eq17}
\end{equation}

where $\bm{\zeta}$ and $\bm{\psi}$  represent the parameters of the entropy model for the quantized latent representaion $\bm{\hat{y}}$ and the hyper-prior $\bm{\hat{z}}$, respectively.

\subsection{\textbf{Overall Pipeline}}

Fig. 1 shows the overall architecture of our proposed framework. The transformation component is a symmetric encoder-decoder, where the encoder consists of multiple Transformer blocks and downsampe units, and the decoder is comprised of several Transformer blocks and upsample units. Given an input image $\bm{x}$, the encoder first applies a downsample module to obtain a new representation with halved spatial size and embedding dimension $C$. Next, the encoder hierarchically reduces the spatial size of the feature map by half while expanding the channel capacity by a factor of two to obtain the compact latent representation $\bm{y}$. The decoder takes the quntized latent representation $\bm{\hat{y}}$ as input and progressively recovers the reconstructed image $\bm{\hat{x}}$.

\subsection{\textbf{ Hybrid Spatial-Channel Attention Transformer Block (HSCATB)}}
Our proposed Transformer block, called the Hybrid Spatial-Channel Attention Transformer Block (HSCATB), is comprised of a Spatial-aware Self-Attention (SaSA) module, Channel-aware Self-Attention (CaSA) module, and a Mixed Local-Global Feed-Forward Network (MLGFFN), which will be explained in the following sections.
\subsubsection{\textbf{Spatial-aware Self-Attention (SaSA)}}
The spatial-aware self-attention mechanism is designed to independently handle high and low frequencies within a feature map at the attention layer. Initially, the given input feature map $\bm{F} \in \mathbb{R}^{H \times W \times C}$ is divided into non-overlapping windows $[\bm{F^1},...,\bm{F^M}]$ where
each window contains $2s \times 2s$ tokens and $M=\frac{H \times W}{2s \times 2s}$. Subsequently, two sets of queries, keys, and values are produced using the following linear projection: 
\begin{align}
\begin{split}
&\bm{Q}^h=[{\bm{F}^1,...,\bm{F}^M]}{{\bm{W}_q}^h}\\
&\bm{Q}^l=[{\text{GAP}(\bm{F}^1),...,\text{GAP}(\bm{F}^M)]}{{\bm{W}_q}^l},
\end{split}
\end{align}
\begin{align}
\begin{split}
&\bm{K}^h=[\bm{F}^1,...,\bm{F}^M]{{\bm{W}_k}^h}\\
&\bm{K}^l=[{\text{GAP}(\bm{F}^1),...,\text{GAP}(\bm{F}^M)]}{\bm{W}_k}^l,
\end{split}
\end{align}
\begin{align}
\begin{split}
&\bm{V}^h=[\bm{F}^1,...,\bm{F}^M]{{\bm{W}_v}^h}\\
&\bm{V}^l=[{\text{GAP}(\bm{F}^1),...,\text{GAP}(\bm{F}^M)]}{\bm{W}_v}^l,
\end{split}
\end{align}
where $\bm{W_q}^h, \bm{W_k}^h, \bm{W_v}^h \in \mathbb{R}^ {C \times C_1}$ and $\bm{W_q}^l, \bm{W_k}^l, \bm{W_v}^l \in \mathbb{R}^ {C \times C_2}$ are learnable projection matrices for high-frequency and low-frequency queries, keys, and values, respectively and GAP denotes an global average pooling function. The obtained $\bm{Q}^h, \bm{K}^h, \bm{V}^h \in \mathbb{R}^ {H \times W \times C_1}$  are in the high-frequency group, similar to a traditional Transformer. In contrast, the resulting  $\bm{Q}^l \in \mathbb{R}^{H \times W \times C_2} $ and $\bm{K}^l, \bm{V}^l  \in \mathbb{R}^{h \times w \times C_2} $ are in low-requency set, where $h=\frac{H}{2s}$ and $w=\frac{W}{2s}$. It should be noted that the sum of the channel numbers of the high-frequency and low-frequency groups is the same as the input feature, indicating that $C_1+C_2=C$.

The resulted low-frequency and high-frequency queries, keys, and values are fed into a window-based multi-head attention (W-MSA), which can generate their output in parallel as below:
\begin{align}
\begin{split}
&\bm{O}^h=\text{W-MSA}(\bm{Q}^h, \bm{K}^h, \bm{V}^h)\\
&\bm{O}^l= \text{W-MSA}(\bm{Q}^l, \bm{K}^l, \bm{V}^l),
\end{split}
\end{align}
In the end, the outputs of these two parallel groups will be fused together to create the final output:
\begin{equation}
\text{SaSA}=\text{Concat}(\bm{O}^h \bm{W}_h, \bm{0}^l \bm{W}_l ), 
\end{equation}

where concat denotes channel concatenation and $\bm{W}_h \in \mathbb{R}^{ C_1 \times C_1} $, $\bm{W}_l \in \mathbb{R}^{ C_2 \times C_2} $ are two learnable projection matrices.

As illustrated in Fig. 2 (a), the high-frequency attention operates within $2s \times 2s $ windows, concentrating on local details within each window. Conversely, the low-frequency attention performs cross-scale attention between query and average-pooled key, value tokens in each window. With global average pooling acting as a low-pass filter, the low-frequency attention captures the global structure of each window, complementing the high-frequency attention's focus on local details. Consequently, these two attention mechanisms exhibit two sizes of receptive fields on both the input, thereby enhancing the capture of short-range and long-range information within the input feature map.

\subsubsection{\textbf{Channel-aware Self-Attention (CaSA)}}
Since SaSA is designed to extract spatial information, including both low-frequency and high-frequency contexts, we include a channel-aware self-attention (CaSA) module in the Transformer block to capture information across channels. As shown in Fig. 2 (b), the information across the spatial dimensions of the feature map is first aggregated. This aggregated information is then passed through a small neural network to generate the attention score, which will be used to weight the input feature map. Given an input feature map, the channel attention mechanism can be formulated as follows:
\begin{equation}
\text{CaSA} = \bm{F}*\text{Sigmoid} (\text{ReLU}( \text{GAP}(\bm{F})\bm{W}_1)\bm{W}_2). 
\end{equation}
Here, $*$ denotes channel-wise multiplication, $\bm{W}_1$ and 
$\bm{W}_2$ $\in \mathbb{R}^ {C \times {C/ \beta} }$ are fully connected layers with a ReLU activation in between. $\beta$ is a channel shrinking factor. GAP stands for global average pooling, and the Sigmoid function constrains the channel attention map to the range .  represents a channel shrinking factor. Channel-aware attention focuses on the most relevant channels, allowing the network to better capture and utilize critical features. This approach helps reduce redundancy and preserve important details.
\subsubsection{\textbf{Mixed Local-Global Feed Forward Network (MLGFFN)}}
The feed-forward network (FFN) is essential within the Transformer block, recognized for its capability to enhance feature representation. We propose a mixed local-global feed-forward network (MLGFFN) to improve the extraction of both local and global information within the Transformer block. In this context, we develop a MLGFFN by incorporating two parallel multi-scale depth-wise convolution paths and one parallel global average pooling path into the transmission process, as depicted in Fig. 3 (d) . MLGFFN divides the input feature map into parts: one part is used in the depth-wise convolution paths, and the other part is employed in the global average pooling path. Given an input feature map $\bm{X}_n$, after layer normalization (LN), the first half of the channels are fed into two parallel branches, where the 3×3 and 5×5 depth-wise convolutions are used to enhance multi-scale local information extraction. Then, the second half of the channels are fed into the global average pooling block to exploit global information. Given an input feature map $\bm{X}_n \in \mathbb{R}^{ H \times W \times C} $, the computation of MLGFFN is formulated as:
\begin{align}
\begin{split}
&\hat{\bm{X}_n}=\text{LN}(\bm{X}_n),\quad [\hat{\bm{X}}^1_n,\hat{\bm{X}}^2_n ]=\text{Split}(\hat{\bm{X}_n}),\\
&\bm{X}^g= \text{GELU}(\text{GAP}( \hat{\bm{X}}^2_n))*\hat{\bm{X}}^2_n,\\
&\bm{X}^l=\text{GELU}({f^{dwc}_{5\times5}}({f^{c}_{1\times1}}(\hat{\bm{X}}^1_n)))*{f^{dwc}_{5\times5}}({f^{c}_{1\times1}}(\hat{\bm{X}}^1_n)),\\
&X_{n+1}=X_n+ \text{Concat}(\bm{X}^g,\bm{X}^l)
\end{split}
\end{align}

\subsection{\textbf{Channel-wise Entropy Model}}
As the entropy model more accurately reflects the true distribution of the latent representation, the bit rate of the compressed file decreases. Minnen \emph{et al.} \cite{minnen2018joint} improved latent distribution prediction by combining a local spatial context with a hyper-prior network. To overcome the issue of slow decoding times, \cite{minnen2020channel} later introduced a channel-wise context model as an alternative to the sequential spatial context model. Given the lack of prioritization among uniformly divided channel groups, \cite{he2022elic} developed a method using unevenly chunked channel slices, which allows lower-entropy channels to depend on higher-entropy ones. We have implemented this unevenly clustered channel-conditional framework for our entropy model. In this model, the latent representation with  $N$ channels, is divided into five chunks along the channel dimension: $16$, $16$, $32$, $64$, $128$ and $M-128$ channels. Each chunk relies on all previously decoded chunks. Assuming that each latent element follows a Gaussian distribution, the parameters of the entropy model can be derived as:
\begin{equation}
(\mu_i,\sigma_i)=epm(g_{uc}(\bm{\hat{y}_{<i}};\bm{{\theta}_{uc}}),h_s(\bm{\hat{z}};\bm{{\theta}_h}),\bm{{\theta}_{ep}}),
\end{equation}

The unevenly channel-wise context model $g_{uc}(.)$ is implemented based on the convolutional mask over the channel dimension. $epm(.)$ shows the entropy parameter function, and $\bm{\hat{y}_{<i}}$ denotes the previously decoded latent elements. $h_s(.)$ demonstrates synthesis transform for the hyper-prior $\bm{\hat{z}}$.

 \subsection{\textbf{Quantization}}
 To support end-to-end training, the quantization process must be replaced with a soft differentiable alternative. In our approach, we approximate hard quantization by adding uniform noise to the latent representation \cite{balle2017endtoend}. This method models the conditional probability of each latent variable as a univariate Gaussian distribution, with its mean and variance convolved with a uniform distribution:
 \begin{equation}
P_{\bm{\hat{y}} | \bm{\hat{z}}}(\bm{\hat{y}}|{\bm{\hat{z}}})=\prod_{i=1}(\mathcal{N}(\mu_{i},\,{\sigma_{i}}^2) \ast \mathcal{U}(-\frac{1}{2},\frac{1}{2}))(\hat{y}_i),
\end{equation}
where the entropy model determines the mean $\mu_{i}$ and variance $\sigma_{i}$. 

\section{\textbf{Experiments}}
 \subsection{\textbf{Implementation Details}}

To train our proposed learned image compression network, we use a combined dataset from Flickr2W \cite{liu2020unified} and ImageNet-1k \cite{deng2009imagenet}. We train the models with a batch size of $16$, where each batch consists of randomly cropped $256\times256$ patches. To obtain a wide range of bitrates, we employ Lagrangian multipliers  $\lambda$  of $\{ {0.0025, 0.0035, 0.0067, 0.0130, 0.0250, 0.0500}\}$ during training. The models are trained using the Adam optimizer \cite{kingma2014adam} over a total of $3.2$ million steps. The initial learning rate is set to $1 \times 10^{-4}$ and is gradually reduced to $10^{-5}$  during the final $0.4$ million steps of training. 

For HSCATB blocks, the base window size $s$ is set to 4 during the nonlinear analysis and synthesis. The channel number $M$ for the latent representation $y$ is set to $320$ and  $C$ is chosen to be $40$. Our proposed model is evaluated on the Kodak image set \cite{kodak}, which includes $24$ images with dimensions of $768 \times 512$ pixels. We utilize   Peak Signal-to-Noise Ratio (PSNR) to assess distortion and bits per pixel (BPP) to measure bitrates.

 \subsection{\textbf{Rate-Distortion performance}}
We compare our proposed model with state-of-the-art neural image compression methods, including Transformer-based \cite{liu2023learned,zafari2023frequency} and convolution-based models \cite{he2022elic,balle2018a}, as well as traditional codecs such as JPEG \cite{jpeg}, JPEG2000 \cite{jpeg2000}, BPG \cite{bellard2015bpg}, and VTM \cite{vtm2022}. The rate-distortion performance on the Kodak dataset is displayed in Fig. 3. Distortion is measured by the Peak Signal-to-Noise Ratio (PSNR). As shown in the Fig. 3, our model achieves better results than all mentioned approaches.

 \begin{figure}
    
  \centering
  \scalebox{0.62}{\includegraphics{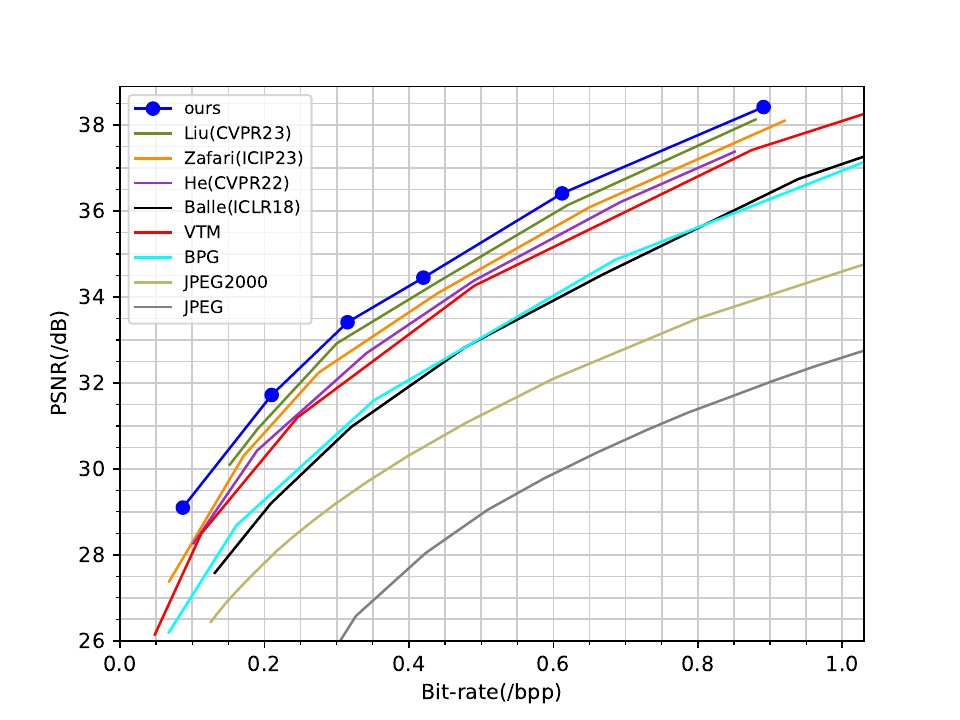}}
  \caption{Rate-distortion performance assessed using the Kodak dataset.}
  \label{fig:High Leval Diagram}
\end{figure}

\subsection{ \textbf{Ablation Study}}
\begin{table}[]
\caption{ Ablation study for different components in HSCATB.  all models are optimized with $\lambda$ = 0.0035}
\scalebox{1.25}{
\begin{tabular}{c c c c c} 
\hline\hline 
\textbf{Model} \vline\vline  & \textbf{LF} \vline \vline & \textbf{HF} \vline\vline&  \textbf{CaSA} \vline\vline & \textbf{PSNR(dB)} \\ [0.2ex] 
\hline 
(a) & \checkmark & \checkmark & \checkmark & \textbf{31.53} \\ 
\hline
(b) &\checkmark & ---   & \checkmark&  30.92 \\
\hline
(c) &---  & \checkmark & \checkmark & 31.06\\
\hline
(d)&\checkmark  & \checkmark &---  & 31.34 \\
\hline \hline 
\end{tabular}}
\label{result}
\centering
\end{table}

\subsubsection{\textbf{Effectiveness of Spatial-aware and Channel-aware Attention in HSCATB}}

Our HSCATB block utilizes attention modules, including spatial-aware self-attention (SaSA) that aggregates low-frequency and high-frequency paths, and channel-aware self-attention (CaSA). To understand the impact of these attention mechanisms, we perform comprehensive ablation studies at approximately consistent bit rates, the results of which are reported in the table. The baseline model (a), which integrates both spatial and channel-wise attention components, achieves the highest PSNR. In contrast, model (b), which incorporates only low-frequency (LF) aggregation in SaSA, experiences an average $0.61$ dB performance drop. Model (c), which relies solely on high-frequency (HF) aggregation in SaSA, shows an average degradation of $0.47$ dB  in performance. These findings suggest that the proposed SaSA effectively captures both high and low-frequency details. Additionally, model (d) without channel-aware self-attention (CaSA) shows an average decrease of $0.19$ dB in performance, indicating that channel-wise global information contributes significantly to compression efficiency.\\

\begin{table}[]
\caption{ Ablation study for different feed forward networks. all models are optimized with $\lambda$ = 0.005}
\scalebox{1.2}{
\begin{tabular}{c c c } 
\hline\hline 
\textbf{Model}   & \textbf{PSNR(dB)}  & \textbf{MS-SSIM}    \\ [0.2ex] 
\hline 
FFN & 33.84 & 0.9607  \\ 
\hline
MLGFFN w/o Local &33.91 & 0.9615   \\
\hline
MLGFFN w/o Global & 34.12  & 0.9624 \\
\hline
MLGFFN & \textbf{34.59}  & \textbf{0.9659} \\
\hline \hline 
\end{tabular}}
\label{result}
\centering
\end{table}

\subsubsection{\textbf{The effectiveness of proposed MLGFFN:} } To illustrate the effectiveness of the mixed local-global feed forward network (MLGFFN), our ablation study compares several models in terms of PSNR and MS-SSIM performance at approximately the same bit rate: the traditional feed forward Network (FFN) composed of MLP layers \cite{vaswani2017attention}, MLGFFN excluding the depth-wise convolution paths (local branch) (referred to as MLGFFN w/o Local), MLGFFN excluding the global average pooling paths (global branch) (referred to as mlgffn w/o Global), and our proposed MLGFFN model. It should be mentioned that all the models are trained with the same hyperparameter, resulting in approximately consistent bit rates. Firstly, MLGFFN demonstrates superior performance compared to FFN. Secondly, the notable decline in performance upon removing either the depth-wise convolution path or the global average pooling path from MLGFFN underscores the importance of integrating both local and global information.

\section{\textbf{Conclusion}}

We designed a novel Transformer-based image compression method that addresses the limitations of traditional CNN and Transformer approaches by incorporating frequency characteristics. Our method leverages a Hybrid Spatial-Channel Transformer Block, which handles high and low frequencies separately and employs a Channel-aware Self-Attention (CaSA) module for cross-channel information capture. Additionally, the inclusion of a Mixed Local-Global Feed Forward Network (MLGFFN) enhances the extraction of rich information. These innovations collectively improve data decorrelation and compression efficiency. Experimental results demonstrate that our framework outperforms state-of-the-art LIC methods in rate-distortion performance, marking a significant advancement in image compression technology.

\bibliographystyle{IEEEtran}
\bibliography{sample}
\end{document}